\pdfoutput=1
\documentclass[11pt]{article}

\usepackage{acl}

\usepackage{times}
\usepackage{latexsym}

\usepackage[T1]{fontenc}

\usepackage[utf8]{inputenc}

\usepackage{microtype}

\usepackage{algorithm}
\usepackage{algorithmic}
\usepackage{amsmath,amssymb}
\usepackage{bm}
\usepackage{booktabs}
\usepackage{bussproofs}
\usepackage{color}
\usepackage{comment}
\usepackage{enumerate}
\usepackage{mathtools}
\usepackage{multirow}
\usepackage{natbib}
\usepackage{stmaryrd}
\usepackage{url}
\usepackage{caption} %
\usepackage{subcaption} %

\newcommand{\vannoord}{\citet{van-noord-etal-2020-character}}
\newcommand{\vannoordp}{\citep{van-noord-etal-2020-character}}
\newcommand{\Vannoord}{\Citet{van-noord-etal-2020-character}}
\newcommand{\bertsem}{BERT\,+\,sem}
\newcommand{\bertsemchar}{BERT\,+\,sem\,+\,char}
\newcommand{\mbertchar}{mBERT\,+\,char}
\newcommand{\twoencchar}{2-enc\,+\,char}
\newcommand{\spacetoken}{\textbar\,\textbar\,\textbar}
\newcommand{\capitaltoken}{\textasciicircum\textasciicircum\textasciicircum}

\title{Does Character-level Information Always Improve \\ DRS-based Semantic Parsing?}

\author{Tomoya Kurosawa \and Hitomi Yanaka \\
        The University of Tokyo\\
        \texttt{\{kurosawa-tomoya, hyanaka\}@is.s.u-tokyo.ac.jp}}

\begin{document}
\maketitle
\begin{abstract}
  Even in the era of massive language models, it has been suggested that character-level representations improve the performance of neural models.
  The state-of-the-art neural semantic parser for Discourse Representation Structures uses character-level representations, improving performance in the four languages (i.e., English, German, Dutch, and Italian) in the Parallel Meaning Bank dataset.
  However, how and why character-level information improves the parser's performance remains unclear.
  This study provides an in-depth analysis of performance changes by order of character sequences.
  In the experiments, we compare F1-scores by shuffling the order and randomizing character sequences after testing the performance of character-level information.
  Our results indicate that incorporating character-level information does not improve the performance in English and German.
  In addition, we find that the parser is not sensitive to correct character order in Dutch.
  Nevertheless, performance improvements are observed when using character-level information.
\end{abstract}

\section{Introduction}

Character-level information is sometimes helpful in grasping the meanings of words for humans.
Previous studies have suggested that character-level information helps to improve the performance of neural models on various NLP tasks~\citep{cherry-etal-2018-revisiting,zhang-etal-2015-character}.
In multilingual NLP systems, character-level information contributes to performance improvements on Named Entity Recognition tasks~\citep{lample-etal-2016-neural,yu-etal-2018-strength} and semantic parsing tasks~\citep{van-noord-etal-2020-character}.
However, due to the black-box nature of neural models, it is still unclear how and why character-level information contributes to model performance.

The rapid developments of neural models have led to a growing interest in investigating the extent to which these models understand natural language.
Recent works have indicated that pre-trained language models are insensitive to word order on permuted English datasets on language understanding tasks~\cite{sinha-etal-2021-masked,sinha-etal-2021-unnatural,pham-etal-2021-order,hessel-schofield-2021-effective}.
Meanwhile, other works have shown controversial results regarding inductive biases for word order~\cite{abdou-etal-2022-word}, especially in different languages~\cite{ravfogel-etal-2019-studying,white-cotterell-2021-examining}.

In this work, we explore the extent to which neural models capture character order.
By focusing on character order rather than word order, we present an in-depth analysis of the capacity of models to capture syntactic structures across languages.
To analyze whether the importance of character order information differs across languages, we investigate multilingual Discourse Representation Structure (DRS; \citet{kamp-reyle-1993-discourse}) parsing models.
\Vannoord{} proposed an encoder-decoder DRS parsing model incorporating character-level representations.
The study concluded that incorporating character-level representations contributes to performance improvements of the model across languages.
However, the underlying mechanism remains unclear.

We examine the influence of character-level information on DRS-based semantic parsing tasks using the state-of-the-art model~\citep{van-noord-etal-2020-character}.
We analyze whether the model is sensitive to the order of character sequences in various units of granularity (i.e., characters, words, and sentences) across the languages.
In addition, we investigate whether the amount of information per character-level token affects the model performance.
Our data will be publicly available at \url{https://github.com/ynklab/character_order_analysis}.

\begin{table*}[t]
  \centering\small
  \begin{tabular}{ll}\toprule
    Sentence &
        Brad Pitt is an actor. \\ \midrule
    Correct order (unigrams) &
        \verb+^^^ b r a d ||| ^^^ p i t t ||| i s ||| a n ||| a c t o r ||| .+ \\ \midrule
    UNI &
        \verb+a a a a a a a a a a a a a a a a a a a a a a a a a+ \\
    SHF (word-level) &
        \verb+d a r ^^^ b ||| t ^^^ p t i ||| i s ||| a n ||| o t c a r ||| .+ \\
    SHF (sentence-level) &
        \verb+c t r r i i . ||| a ||| d a t ||| b p ||| s t ||| ^^^ o n a ^^^+ \\
    RND &
        \verb+" i c v , t 9 d j : l ' n 6 0 b 0 1 q w ! j w u q+ \\ \midrule
    Bigrams &
        \verb+^^^b br ra ad d||| |||^^^ ^^^p pi it tt t||| |||i is s|||+ \\
    & \verb+|||a an n||| |||a ac ct to or r||| |||.+ \\ \bottomrule
\end{tabular}
  \caption{All of character-level information of the same input sentence \textit{Brad Pitt is an actor}.
  ``\capitaltoken{}'' and ``\spacetoken{}'' are special characters representing capitals and spaces, respectively.}
  \label{table:chars}
\end{table*}

\section{Background}
\paragraph*{Multilingual DRS corpus}
The Parallel Meaning Bank (PMB; \citet{abzianidze-etal-2017-parallel}) is a multilingual corpus annotated with DRSs.
The PMB contains sentences for four languages (English, German, Dutch, and Italian) with three levels of DRS annotation: gold
(fully manually checked), silver (partially manually corrected), and bronze (without manual correction).
The PMB also provides semantic tags, which are linguistic annotations for producing DRSs~\cite{abzianidze-bos-2017-towards-revised}.

\paragraph*{Neural DRS parsing models}
There have been various attempts to improve the performance of neural DRS parsing models, such as by using graph formats~\cite{fancellu-etal-2019-semantic,poelman-etal-2022-transparent}, stack LSTMs~\cite{evang-2019-transition}, and sequence labeling models~\cite{shen-evang-2022-drs}.
\Vannoord{} proposed a sequence-to-sequence model with neural encoders and an attention mechanism \citep{vaswani-etal-2017-attention}.
In the study, the number and type of encoders and the type of embeddings of the pre-trained language models, including BERT~\citep{devlin-etal-2019-bert}, were changed to evaluate the model.
Moreover, linguistic features and character-level representations were added to the model, concluding that character-level representations contribute to the performance improvements in all four languages, compared to using only BERT embeddings as input.

\paragraph*{Sensitivity to word order}
Several studies have analyzed whether generic language models understand word order~\cite{sinha-etal-2021-masked,sinha-etal-2021-unnatural,pham-etal-2021-order,hessel-schofield-2021-effective,abdou-etal-2022-word}.
However, these studies have focused on text classification benchmarks, such as GLUE~\cite{wang2018glue}, rather than semantic parsing tasks, such as DRS parsing.
In addition, these studies did not investigate whether models are sensitive to character order.

\section{Experimental Setup}
We explore whether character-level information influences the predictions of the state-of-the-art DRS parsing model using character representations~\citep{van-noord-etal-2020-character} across languages.
This section introduces the common experimental setup.

\paragraph*{Dataset}
In all experiments, we use the PMB release 3.0.0 and follow the same setup as in the original study~\citep{van-noord-etal-2020-character}.
We use gold test sets for evaluation after fine-tuning.
See Appendix \ref{appendix:pmb} for details of the dataset settings.

\paragraph*{Models}
We focus on two types of architectures: English BERT with semantic tags (\bertsem{}) for English and multilingual BERT (mBERT) for the other languages, achieving the highest F1-scores on the PMB release 3.0.0 in the original study \vannoordp{}.
These setups use a single bi-LSTM encoder for BERT (or mBERT) embeddings and semantic tags (only English), in the previous study. %
Whereas the original model used their trigram-based tagger and predicted semantic tags for English, we use the gold semantic tags in the PMB to exclude performance changes based on the accuracy of the tagger.
Although PMB also has gold semantic tags for non-English languages, we adopt them only for English to compare with \vannoord{}.
We define \bertsemchar{} for English and \mbertchar{} for the other languages with an additional bi-LSTM encoder for character-level representations as the default setting \twoencchar{}.

\paragraph*{Evaluation metrics}
To evaluate model performance precisely, we report averaged micro F1-scores of 15 runs,
which are more than those on the settings of the original study (five runs).
We use Counter and Referee \citep{van-noord-etal-2018-evaluating,van-noord-etal-2018-exploring} to calculate the micro F1-score.
See Appendix \ref{section:drs-eval} for further details.

\begin{figure*}
  \begin{minipage}[b]{0.277\linewidth}
    \centering
    \includegraphics[trim={1mm 1mm 1mm 1mm},clip,width=\hsize]{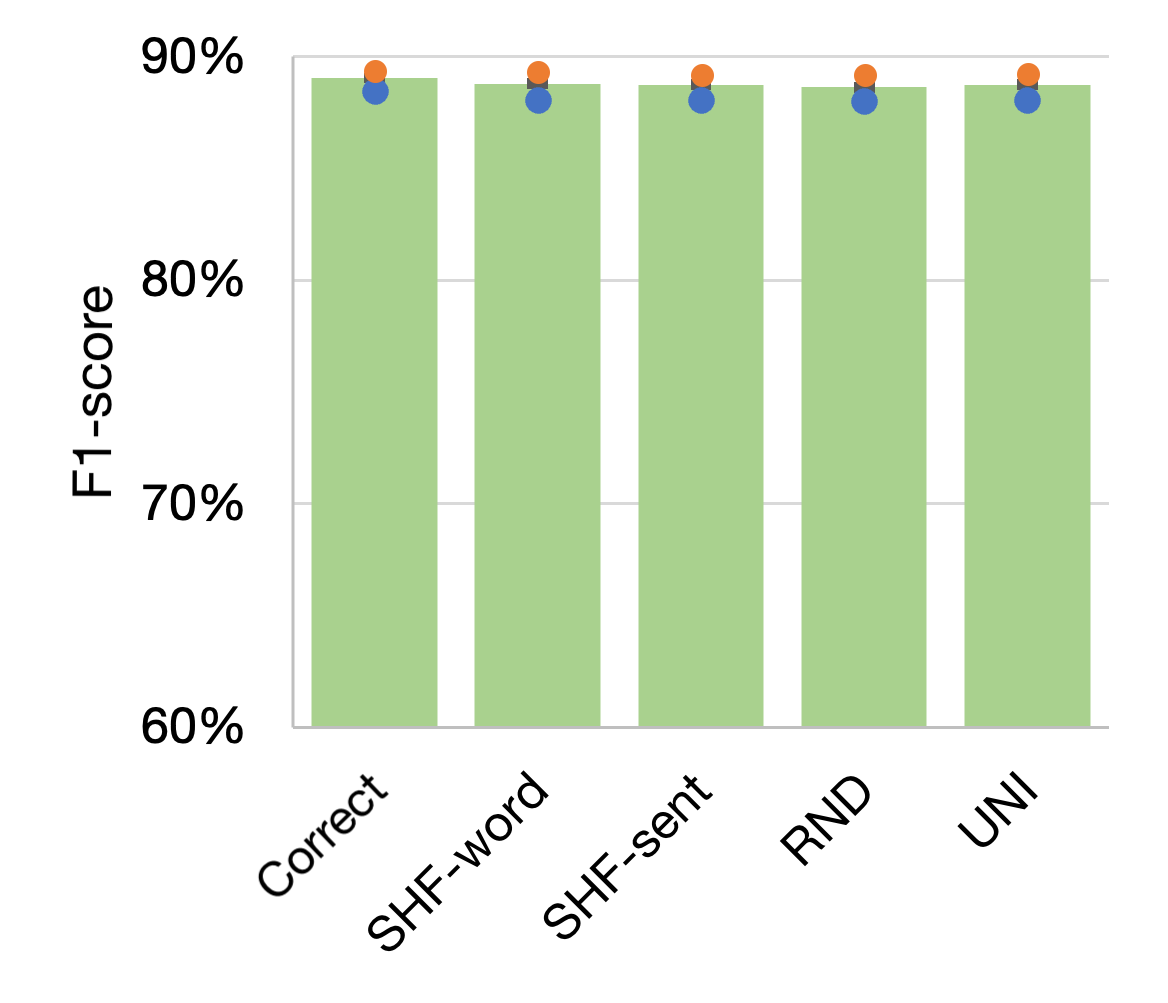}
    \subcaption{English}
    \label{subfigure:english}
  \end{minipage}
  \begin{minipage}[b]{0.226\linewidth}
    \centering
    \includegraphics[trim={1mm 1mm 1mm 1mm},clip,width=\hsize]{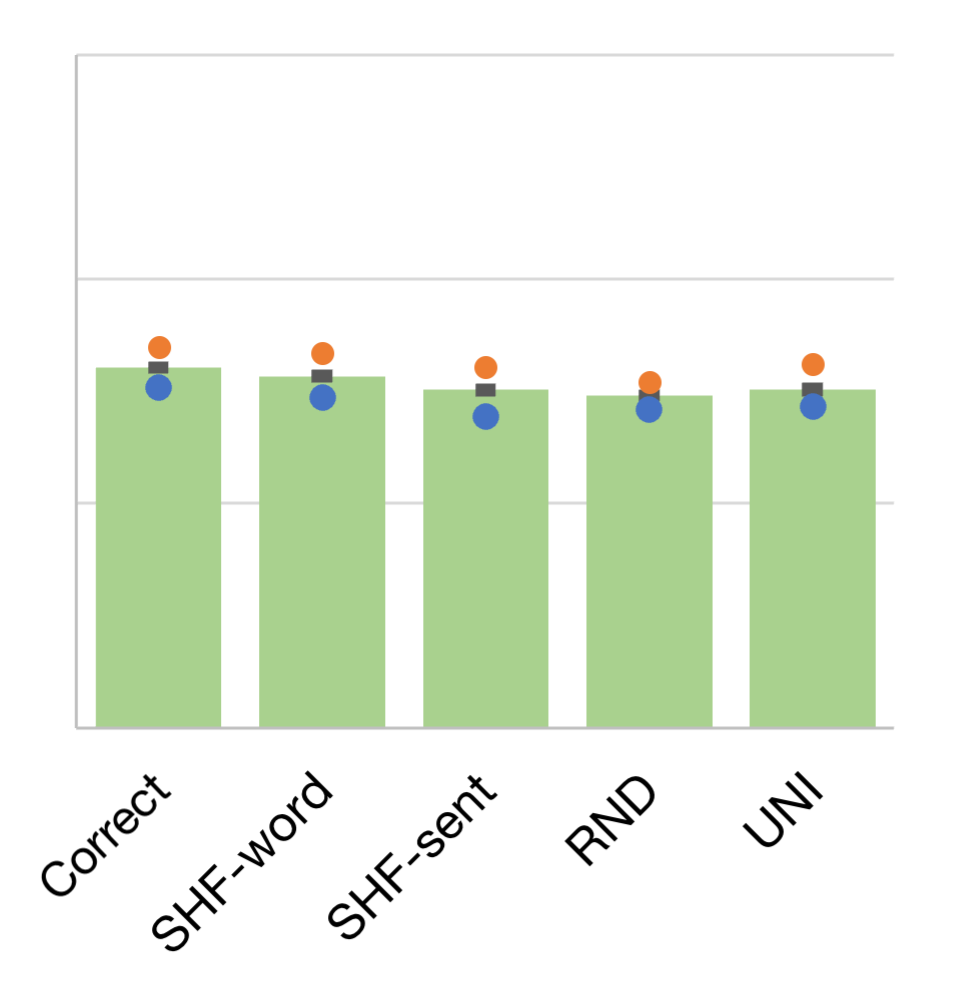}
    \subcaption{German}
    \label{subfigure:german}
  \end{minipage}
  \begin{minipage}[b]{0.226\linewidth}
    \centering
    \includegraphics[trim={1mm 1mm 1mm 1mm},clip,width=\hsize]{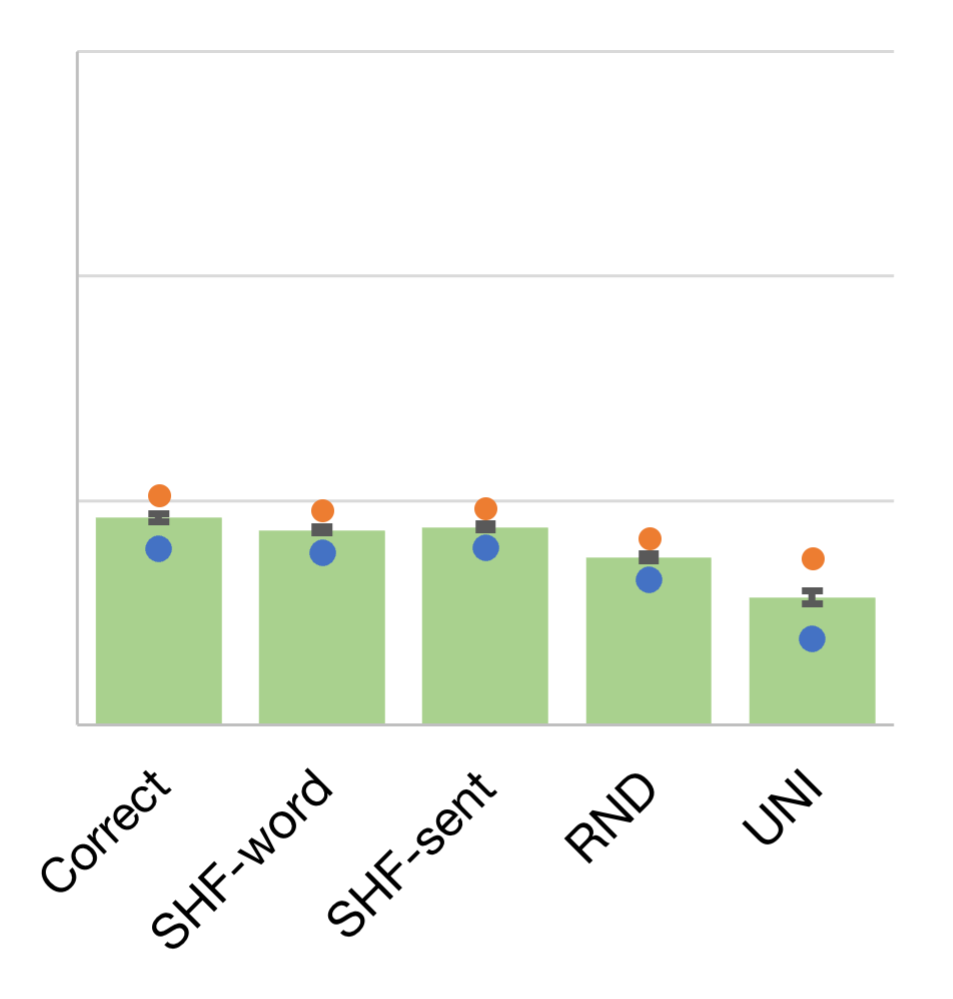}
    \subcaption{Dutch}
    \label{subfigure:dutch}
  \end{minipage}
  \begin{minipage}[b]{0.252\linewidth}
    \centering
    \includegraphics[trim={1mm 1mm 1mm 1mm},clip,width=\hsize]{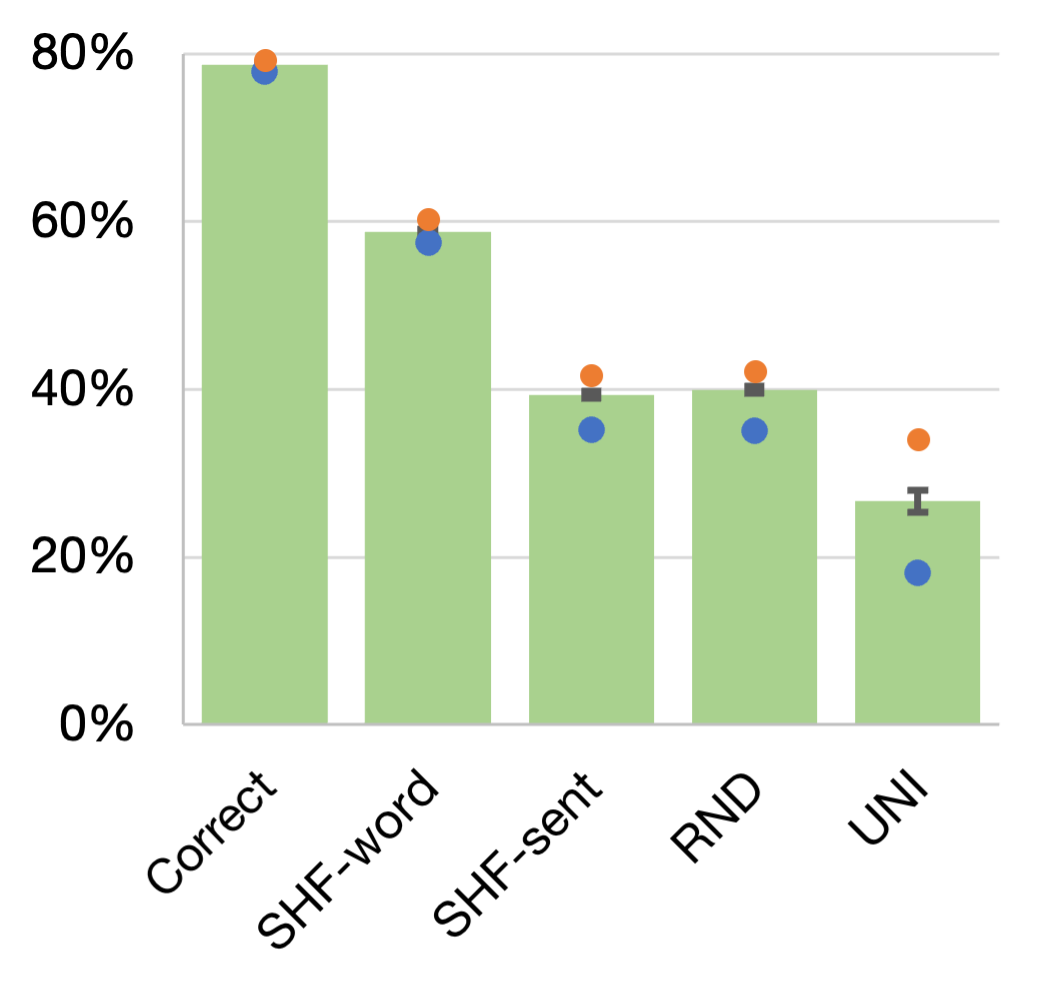}
    \subcaption{Italian}
    \label{subfigure:italian}
  \end{minipage}
  \caption{F1-scores for four languages.
  Green bars show the average scores of runs, including standard error, and blue and orange dots show the minimum and maximum scores, respectively. The exact results are in Appendix \ref{appendix:num-values}.}
  \label{figure:main-result}
\end{figure*}

\section{Method}
We provide multiple methods to \textit{reanalyze} whether the DRS parsing models \vannoord{} are sensitive to character-level information across languages in a more fine-grained way.
First, we \textit{reexamine} whether character-level information benefits the model in terms of character sequences compared to the setup without an encoder for characters.
Second, we examine whether the model trained with correct character order predicts correct DRSs even with incorrect character sequences obtained using techniques such as shuffling.
In the above two methods, we prepare models trained with correct character sequences and evaluate the performance when incorrect character order is input to them.
Third, we explore the capacity of the models to understand character-level information using unigrams or bigrams of characters as character tokens.
By using unigrams, we mean one character at a time, and by using bigrams, we mean two characters at a time.

\subsection{Do models use characters as a clue?}
\label{section:method-char}
Before examining whether the model is sensitive to character order,
we have to reveal whether incorporating character sequences is useful or not for the model.
To test this, we prepare the models trained on correct character order and evaluate them using unified character sequences (UNI).
Note that our method is a more detailed analysis of \vannoord{} in claiming whether character-level information is useful (or not).
UNI consists of a single character \texttt{a} (see Table \ref{table:chars}).
As this type of sequences is entirely irrelevant to the input sentences,
the model should perform almost the same as setups without an encoder for character-level information.
Additionally, we reproduce to compare the values of the no char setups.

\subsection{Are models sensitive to character order?}
\label{section:method-order}
For languages in which the usefulness of character-level information is confirmed (Section \ref{section:method-char}),
we analyze whether the model understands correct character order across languages.
We create two types of incorrect character sequences by (i) shuffling the order of the character sequences and (ii) randomizing the sequences (see Table \ref{table:chars}).
If the model is sensitive to correct character order during training, it should fail to predict the correct DRSs with incorrect order.

\paragraph*{Shuffled (SHF)}
We shuffle the sequences on two levels, word-level and sentence-level.
A word-level shuffled character sequence is obtained by shuffling character order within each word (separated by ``\spacetoken{}'', see Table \ref{table:chars}).
In contrast, a sentence-level shuffled sequence can be created by rearranging the characters in the entire sentence, including spaces.
By comparing the performance of these two shuffling levels, we investigate the extent to which the model is confused, depending on the extent of disturbance in the character order.

\paragraph*{Randomized (RND)} %
We provide an additional types of character sequences, randomized character sequences.
The randomized sequences consist of characters
randomly selected from the PMB in each language.

\subsection{Can models be improved performance by extended character sequences?}
\label{section:method-bigram}
The original model uses a unigram character as the character token.
Typically, the amount of information per character-level token is increased by using bigrams instead of unigrams.
Also, the four languages in the PMB consist of alphabets, and the number of letters is limited, unlike several Asian languages such as Chinese and Japanese.
Thus we provide bigram sequences other than unigram sequences, treated them as extended character sequences, and train the models using them.
In the bigram sequence settings \textsc{(bigrams)}, as illustrated in the bottom line of Table~\ref{table:chars}, the models can obtain not only character order but also the connections of characters from character tokens.
If an encoder for character-level representations affects the model performance, the use of bigram sequences is expected to improve the model performance.

\section{Results and Discussion}
\label{section:result}

\paragraph{Character contribution for models}
Table \ref{table:uni} shows the micro averaged F1-scores with their standard errors.
The values in the \textsc{no char} column are F1-scores of the setups without character encoders.
The stander errors corresponding to English and German showed significant differences.
However, these differences suggest that character-level information is not crucial in DRS parsing.
On the other hand, we can see effectiveness in the other languages: Dutch and Italian.
In particular, an F1-score change of more than 50\% can be observed in Italian.
However, values of \textsc{UNI} are far lower than ones of \textsc{no char} in Dutch and Italian.
This tendency suggests that providing incorrect character-level information decreases scores critically when incorporating character-level information is effective.

\paragraph{Models' sensitivity to character order}
Figure \ref{figure:main-result} shows the micro averaged, maximum, and minimum F1-scores for each type of character-level information: \textsc{correct}, \textsc{SHF-word} (word-level SHF), \textsc{SHF-sent} (sentence-level SHF), \textsc{RND}, and \textsc{UNI} (for comparison).
In English (Figure \ref{subfigure:english}) and German (Figure \ref{subfigure:german}),
only minor changes (1\%) were observed in the averaged F1-scores for all types of characters.
This observation supports less effectiveness of incorporating character-level information for these two languages.
We also experimented with the \twoencchar{} model without semantic tags in English and obtained similar trends (see Appendix \ref{appendix:no-sem}).

In Dutch (Figure \ref{subfigure:dutch}), even though we can see a slight performance decrease from \textsc{correct} to \textsc{RND}, shuffling the character order does not affect the performance of the models.
These results indicate that DRS parsing models are not sensitive to character order for Dutch.

For Italian (Figure \ref{subfigure:italian}), we can see that the correct character order contributes to the performance of the model.
Shuffling the characters within each word decreased the model's performance by 20\% (from 79\% to 59\%).
The performance decreased by another 20\% (from 59\% to 39\%) when shuffling in a whole sentence, compared with \textsc{SHF-word}.
One of the possible reasons that the Italian model is significantly sensitive to the character-level information is the existence of the accented characters specific to Italian (e.g., \'e), especially the loss of it by shuffling characters within sentences (\textsc{SHF-word} $\to$ \textsc{SHF-sent}).
For example, the character \'e plays the role of an auxiliary verb in Italian by itself.
When characters are lost by shuffling them within words (\textsc{correct} $\to$ \textsc{SHF-word}), shuffled character sequences within words appear to affect the incorrect prediction of words.
Further investigation into differences between languages is needed, which is left as future work.

\begin{table}
  \small\centering
  \begin{tabular}{lccc} \toprule
        & \textsc{Correct} & \textsc{UNI} & \textsc{No char} \\ \midrule
English & 89.05 $\pm$ 0.06 & 88.76 $\pm$ 0.09 & 88.89 $\pm$ 0.08 \\
German  & 76.07 $\pm$ 0.12 & 75.09 $\pm$ 0.17 & 75.33 $\pm$ 0.14 \\
Dutch   & \textbf{69.23 $\pm$ 0.18} & \textbf{65.69 $\pm$ 0.30} & \textbf{68.81 $\pm$ 0.13} \\
Italian & \textbf{78.75 $\pm$ 0.10} & \textbf{26.66 $\pm$ 1.30} & \textbf{77.54 $\pm$ 0.09} \\\bottomrule
\end{tabular}
  \caption{F1-scores (\%) on the gold test set depending on character-level information: \textsc{correct} and \textsc{UNI}.}
  \label{table:uni}
\end{table}

\begin{table}
  \small\centering
  \begin{tabular}{lccc} \toprule
        & \textsc{No char} & \textsc{Unigrams} & \textsc{Bigrams} \\ \midrule
English & 88.89 $\pm$ 0.08 & 88.99 $\pm$ 0.08 & 89.10 $\pm$ 0.07 \\
German  & \textbf{75.33 $\pm$ 0.14} & \textbf{75.94 $\pm$ 0.11} & \textbf{76.96 $\pm$ 0.11} \\
Dutch   & \textbf{68.81 $\pm$ 0.13} & \textbf{69.22 $\pm$ 0.18} & \textbf{69.62 $\pm$ 0.11} \\
Italian & \textbf{77.54 $\pm$ 0.09} & \textbf{78.73 $\pm$ 0.11} & \textbf{79.46 $\pm$ 0.08} \\ \bottomrule
\end{tabular}
  \caption{F1-scores (\%) on the gold test set depending on character-level information: \textsc{unigrams} and \textsc{bigrams}.}
  \label{table:bigrams}
\end{table}

\paragraph{Extending character tokens improves model performance}
Table \ref{table:bigrams} shows the averaged F1-scores and standard errors obtained using character-level information (\textsc{bigrams}, \textsc{unigrams}, and \textsc{no char}).
We observe no significant differences in the overall setups in English.
In contrast, in German, Dutch, and Italian, we can find performance improvements in extensions from unigrams to bigrams and from no character-level information to unigrams.
In particular, the model achieves the largest improvements by incorporating unigrams as character-level information in Italian and by extending from unigrams to bigrams in German, respectively.
These results indicate that although models are not usually sensitive to character order, character-level information helps performance improvements in German, Dutch, and Italian.

One of the reasons models cannot achieve any improvements in English, while improvements are observed in non-English languages, is the quantity and quality of data in the PMB.
As noted in the statistics of PMB 3.0.0 (Appendix \ref{appendix:pmb} and Table \ref{table:pmb}), we can use over 6.6k English gold training data.
In addition, nearly 100k sliver cases are available.
In contrast, the German dataset only contains 1.2k gold and 5.3k silver cases, and there is no gold case in both Dutch and Italian.

\section{Conclusion and Future Work}
In this study, we carried out a further exploration of the extent to which character-level representations contribute to the performance improvements of multilingual DRS parsing models.
We found that character-level information provided little performance improvement in English and German but improved performance in Dutch and Italian.
However, we find that the model is sensitive to character order in Italian but not in Dutch.
The take-away message from our investigation is that the importance of character-level information in DRS-based semantic parsing depends on the language and syntactic structures of the sentences.

In future work, we will analyze in more detail the significant differences between the four languages, especially Italian, and other languages.
Another direction of our future work is to investigate the relationship between the neural models and humans in reading performance for incorrect character order.
It would be interesting to analyze whether the results on DRS parsing tasks are consistent with those of these studies \citep{ferreira-etal-2002-good,gibson-etal-2013-rational,traxler-2014-trends}.

\section*{Limitations}
In this study, we focus on DRS parsing tasks, and do not consider other representation formats for semantic parsing tasks.

\section*{Acknowledgements}
We thank the three anonymous reviewers for their helpful comments and suggestions, which improved this paper.
We also thank our colleagues, Aman Jain and Anirudh Reddy Kondapally, for proofreading and providing many comments on our paper.
This work was supported by JST, PRESTO grant number JPMJPR21C8, Japan.

\bibliography{anthology,custom}
\bibliographystyle{acl_natbib}

\clearpage
\appendix
\section{DRS Parsing Task}

DRS parsing is a task to convert natural language sentences into DRS-based meaning representations.
In \vannoord{} and this study, the outputs of the models are clausal forms with relative naming for the variables. %
See \citet{van-noord-etal-2018-exploring} for the further details.

\subsection{Evaluation}
\label{section:drs-eval}
This study follows micro F1-scores based on matching clauses between predicted and gold DRSs adopted by \vannoord{}.
The tool for calculating the values is Counter \citep{van-noord-etal-2018-evaluating}, which searches for the best mapping of variables between two DRSs and calculates the values based on the number of clauses.
Referee \citep{van-noord-etal-2018-exploring} verifies whether an output DRS is well-formed.
An output DRS is ill-formed (i.e., not well-formed) when it has illegal clauses or the tool fails to solve variable references.

\section{Dataset Settings}
We use PMB release 3.0.0 and the same setup as that in the previous study~\citep{van-noord-etal-2020-character}.
As pre-training datasets, we use a merged set of the gold and the silver training sets for English, a merged set of all training sets (gold, silver, and bronze) for German\footnote{We also experiment on the setup described in \vannoord{}. See Appendix \ref{section:vannoord-german}}, and combined sets of silver and bronze training sets for Dutch and Italian.
As datasets for fine-tuning, we use the gold training set for English, a combined set of the gold and silver training sets for German, and the silver training sets for Dutch and Italian.
\label{appendix:pmb}
Table \ref{table:pmb} shows data statistics of the PMB release 3.0.0.

\begin{table}[t]
  \centering\small
  \begin{tabular}{l|ccc|c|c} \toprule
        & & \textbf{Gold} & & \textbf{Silver} & \textbf{Bronze} \\
 & Train     & Dev       & Test      & Train           & Train           \\ \midrule
English   & 6,620     & 885       & 898       & 97,598          & 146,371         \\
German    & 1,159     & 417       & 403       & ~~5,250          & 121,111         \\
Dutch     & ~~~~~~~0     & 529       & 483       & ~~1,301          & \,~21,550         \\
Italian   & ~~~~~~~0     & 515       & 547       & ~~2,772          & \,~64,305         \\\bottomrule
\end{tabular}
  \caption{The data statistics of PMB release 3.0.0.}
  \label{table:pmb}
\end{table}

\section{Numerical Results}
\label{appendix:num-values}
Table \ref{table:numerical-value} shows numerical values reported in Figure \ref{figure:main-result}.

\begin{table*}[t]
  \vspace{5\baselineskip}
  \begin{minipage}[t]{\linewidth}
    \centering\small
    \begin{tabular}{lccccc} \toprule
         & Avg & SE & Min & Max & Avg values per pre-train\\ \midrule
\textsc{Correct}  & 89.05   & 0.06     & 88.47   & 89.39 & 89.04,\, 88.95,\, 89.17  \\
\textsc{SHF-word} & 88.80   & 0.09     & 88.03   & 89.34 & 88.80,\, 88.75,\, 88.87  \\
\textsc{SHF-sent} & 88.75   & 0.09     & 88.04   & 89.20 & 88.79,\, 88.53,\, 88.93  \\
\textsc{RND}      & 88.65   & 0.09     & 88.01   & 89.19 & 88.74,\, 88.48,\, 88.74  \\
\textsc{UNI}      & 88.76   & 0.09     & 88.04   & 89.25 & 88.62,\, 88.61,\, 89.05  \\\bottomrule
\end{tabular}
    \subcaption{English\\~}
  \end{minipage}
  \vspace{.5\baselineskip}
  \begin{minipage}[t]{\linewidth}
    \centering\small
    \begin{tabular}{lccccc} \toprule
         & Avg & SE & Min & Max & Avg values per pre-train\\ \midrule
\textsc{Correct} & 76.07   & 0.12     & 75.21   & 77.02 & 76.24,\, 76.24,\, 75.74 \\
\textsc{SHF-word} & 75.68   & 0.13     & 74.76   & 76.75 & 75.69,\, 75.88,\, 75.46 \\
\textsc{SHF-sent} & 75.07   & 0.13     & 73.90   & 76.09 & 74.89,\, 75.28,\, 75.03 \\
\textsc{RND}      & 74.81   & 0.11     & 74.22   & 75.46 & 74.72,\, 74.83,\, 74.88 \\ 
\textsc{UNI}      & 75.09   & 0.17     & 74.34   & 76.26 & 75.02,\, 75.25,\, 74.99 \\\bottomrule
\end{tabular}
    \subcaption{German}
  \end{minipage}
  \vspace{.5\baselineskip}
  \begin{minipage}[t]{\linewidth}
    \centering\small
    \begin{tabular}{lccccc} \toprule
         & Avg & SE & Min & Max & Avg values per pre-train \\ \midrule
\textsc{Correct} & 69.23   & 0.18     & 67.89   & 70.26 & 69.41,\, 69.33,\, 68.95  \\
\textsc{SHF-word} & 68.69   & 0.13     & 67.70   & 69.60 & 68.94,\, 68.68,\, 68.46  \\
\textsc{SHF-sent} & 68.82   & 0.13     & 67.95   & 69.68 & 69.31,\, 68.59,\, 68.55  \\
\textsc{RND}      & 67.47   & 0.14     & 66.52   & 68.34 & 67.65,\, 67.50,\, 67.26  \\
\textsc{UNI}      & 65.69   & 0.30     & 63.90   & 67.47 & 65.68,\, 65.76,\, 65.64  \\ \bottomrule
\end{tabular}
    \subcaption{Dutch}
  \end{minipage}
  \vspace{.5\baselineskip}
  \begin{minipage}[t]{\linewidth}
    \centering\small
    \begin{tabular}{lccccc} \toprule
         & Avg & SE & Min & Max & Avg values per pre-train \\ \midrule
\textsc{Correct} & 78.75   & 0.10     & 77.99   & 79.29 & 78.97,\, 78.53,\, 78.75 \\
\textsc{SHF-word} & 58.84   & 0.20     & 57.59   & 60.30 & 58.74,\, 58.34,\, 59.43 \\
\textsc{SHF-sent} & 39.37   & 0.42     & 35.22   & 41.78 & 39.37,\, 38.08,\, 40.66 \\
\textsc{RND}      & 39.95   & 0.46     & 35.08   & 42.26 & 39.83,\, 40.30,\, 39.73 \\ 
\textsc{UNI}      & 26.66   & 1.30     & 18.23   & 34.16 & 28.06,\, 30.14,\, 21.77 \\\bottomrule
\end{tabular}
    \subcaption{Italian}
  \end{minipage}
  \caption{The numerical values (\%) reported in Figure \ref{figure:main-result}. SE is the abbreviation of standard error.}
  \label{table:numerical-value}
\end{table*}

\section{Results in English without Semantic Tags}
\label{appendix:no-sem}
\begin{figure*}[t]
  \centering
  \includegraphics[trim={1mm 1mm 1mm 1mm},clip,width=.25\hsize]{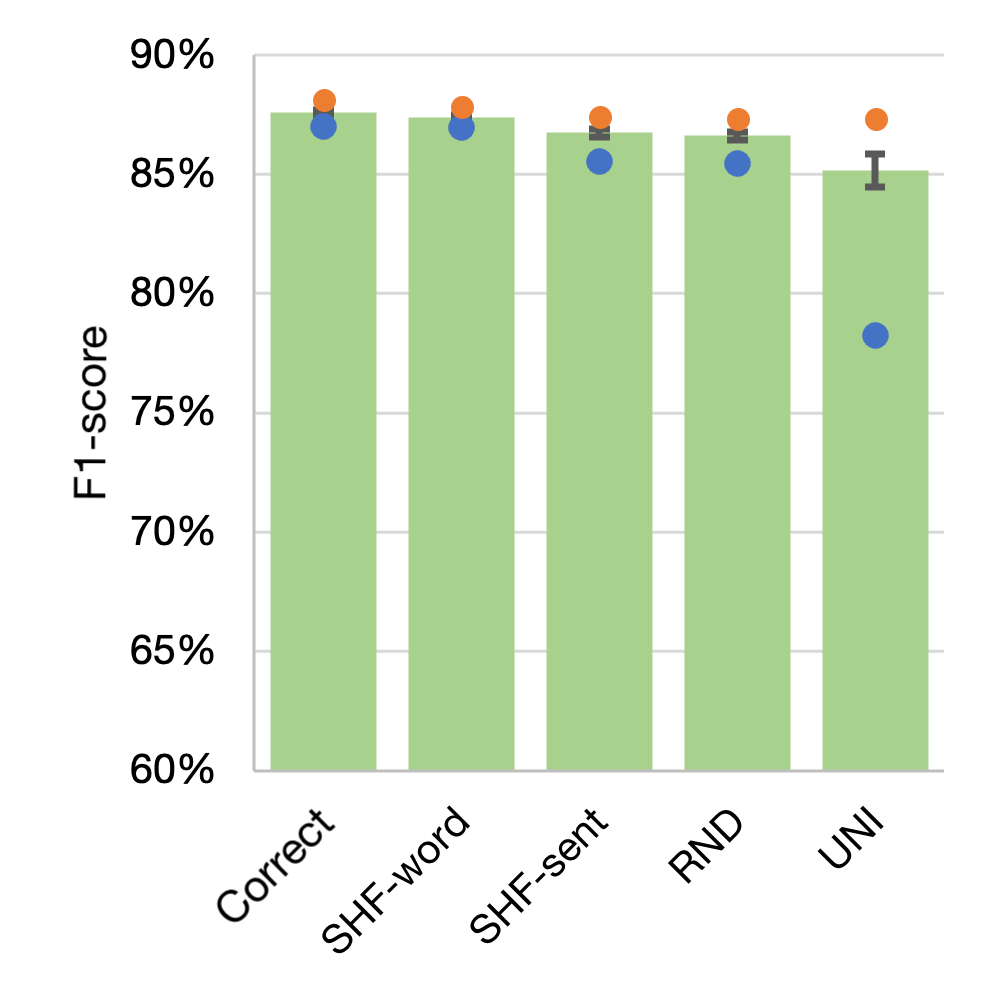}
  \caption{F1-scores of the gold test set predicted by the \twoencchar{} model without semantic tags in English.}
    \label{figure:no-sem}
\end{figure*}

Figure \ref{figure:no-sem} and Table \ref{table:no-sem} show the results of the \twoencchar{} model without semantic tags in English.
Compared with \twoencchar{} (Figure \ref{subfigure:english}), we can observe slightly larger but minor changes in the averaged F1-scores.
Thus, regardless of the existence of semantic tags, our experimental results indicate that the model is not sensitive to the order of character sequences in English.

\begin{table*}[t]
    \centering\small
    \begin{tabular}{lcccc} \toprule
         & Avg & SE & Min & Max \\ \midrule
\textsc{Correct} & 87.58   & 0.10     & 87.01   & 88.14   \\
\textsc{SHF-word} & 87.39   & 0.08     & 86.97   & 87.84   \\ 
\textsc{SHF-sent} & 86.73   & 0.16     & 85.54   & 87.40   \\
\textsc{RND}      & 86.61   & 0.17     & 85.48   & 87.34   \\
\textsc{UNI}      & 85.15   & 0.70     & 78.25   & 87.34   \\ \bottomrule
\end{tabular}
    \caption{The numerical values (\%) reported in Figure \ref{figure:no-sem}, the \twoencchar{} model without semantic tags in English. SE is the abbreviation of standard error.}
    \label{table:no-sem}
\end{table*}

\section{Additional Analysis}

\subsection{Score change by character-level information per case}
\begin{figure*}
  \begin{minipage}[b]{0.291\linewidth}
    \centering
    \includegraphics[trim={1mm 1mm 1mm 1mm},clip,width=.95\hsize]{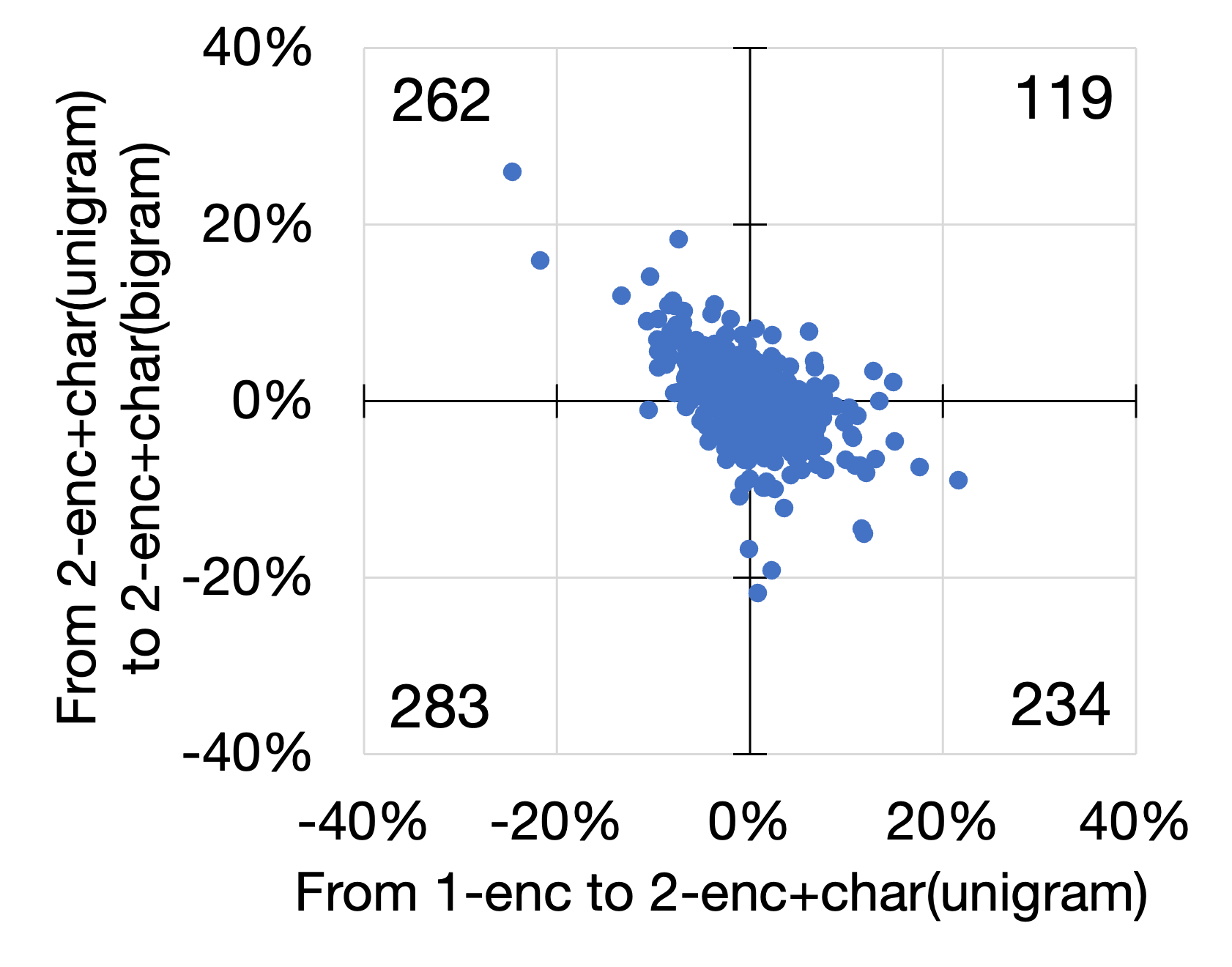}
    \subcaption{English (898 cases)}
    \label{subfigure:indiv-english}
  \end{minipage}
  \begin{minipage}[b]{0.23\linewidth}
    \centering
    \includegraphics[trim={1mm 1mm 1mm 1mm},clip,width=.95\hsize]{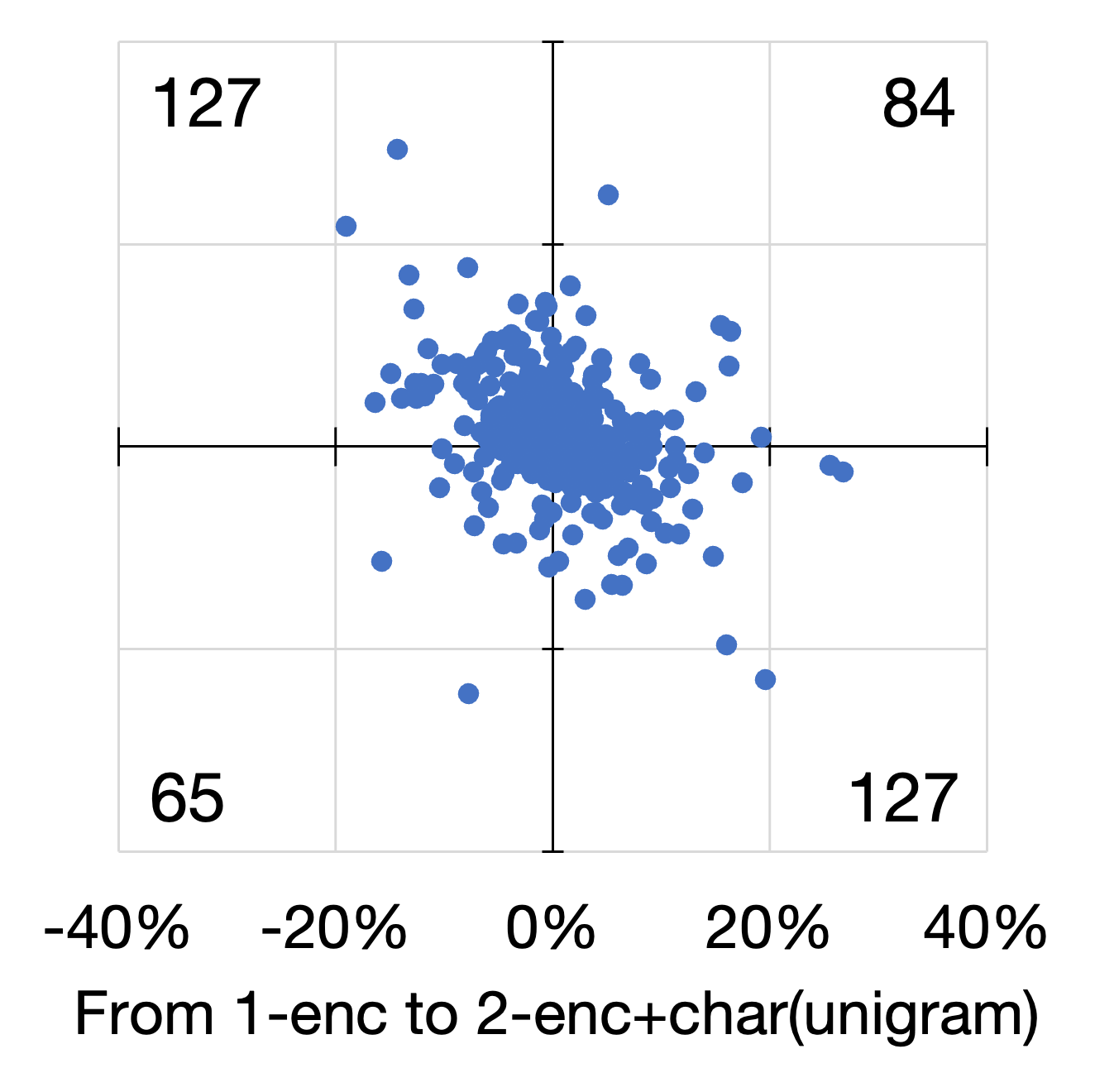}
    \subcaption{German (403 cases)}
    \label{subfigure:indiv-german}
  \end{minipage}
  \begin{minipage}[b]{0.23\linewidth}
    \centering
    \includegraphics[trim={1mm 1mm 1mm 1mm},clip,width=.95\hsize]{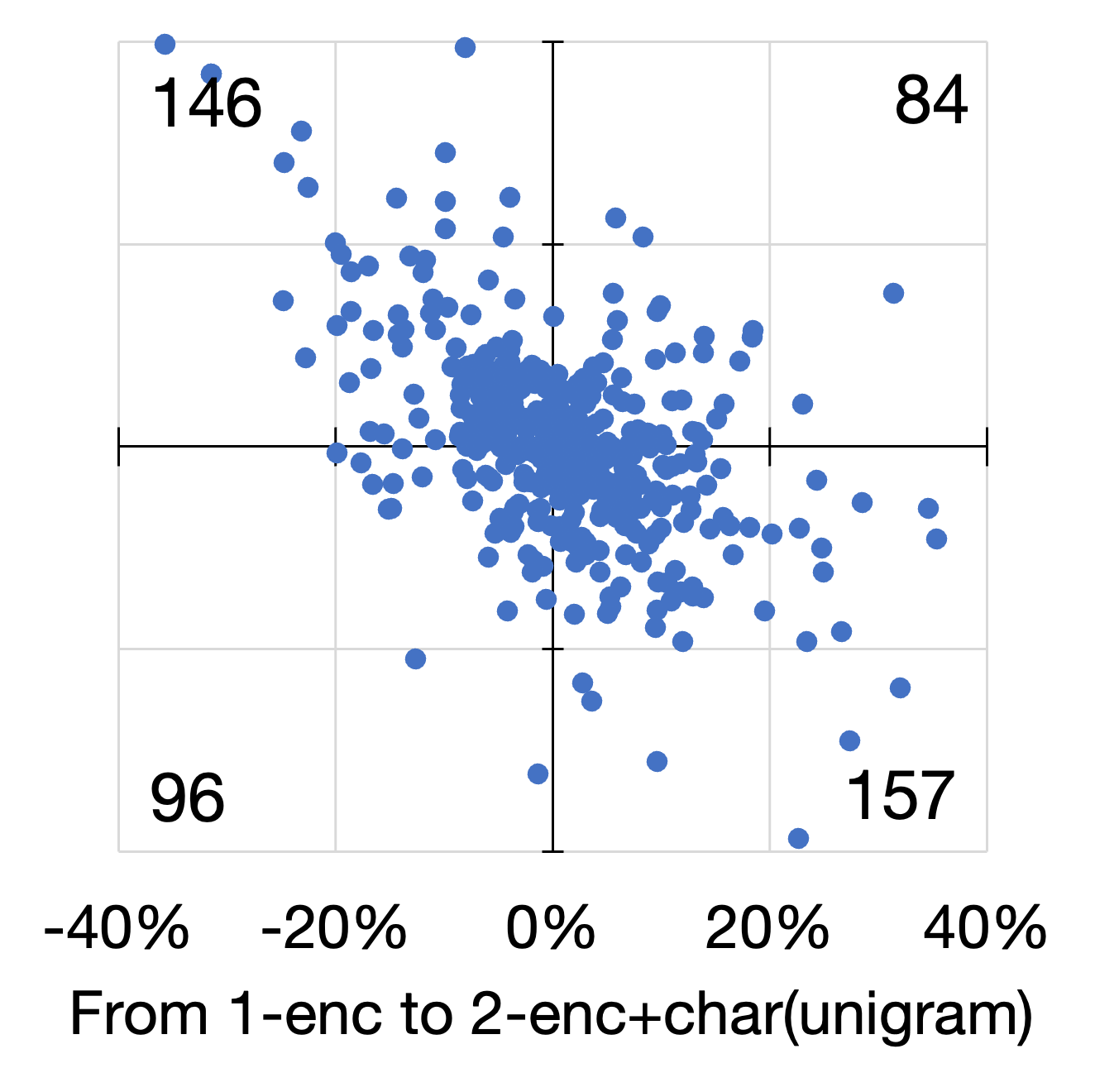}
    \subcaption{Dutch (583 cases)}
    \label{subfigure:indiv-dutch}
  \end{minipage}
  \begin{minipage}[b]{0.23\linewidth}
    \centering
    \includegraphics[trim={1mm 1mm 1mm 1mm},clip,width=.95\hsize]{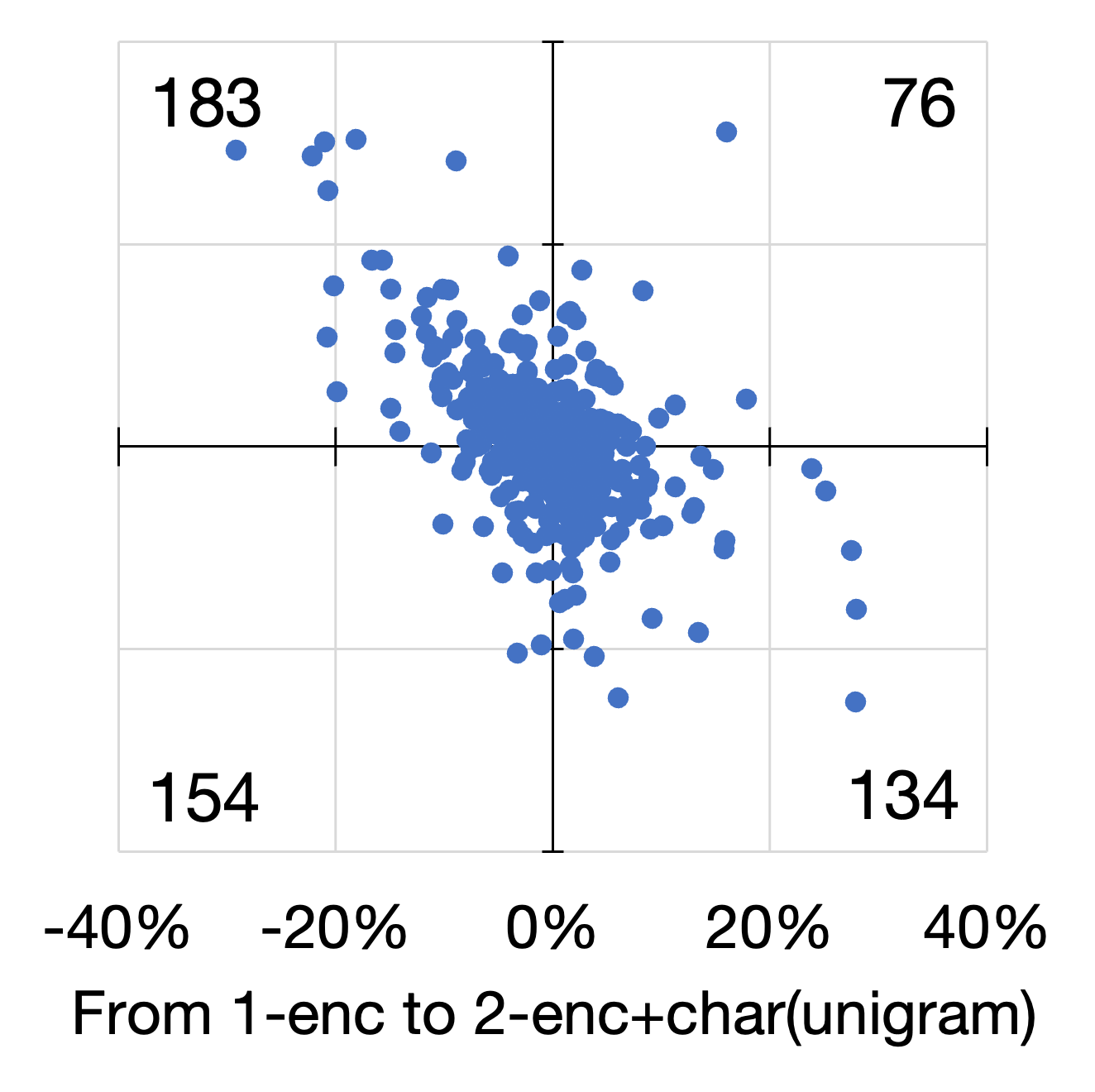}
    \subcaption{Italian (547 cases)}
    \label{subfigure:indiv-italian}
  \end{minipage}
  \caption{Distribution of F1-score changes from \textsc{no char} to \textsc{unigrams} (x-axis) and from \textsc{unigrams} to \textsc{bigrams} (y-axis) per case on the gold test set of the four languages.
  The numbers on the corners are the numbers of cases in each quadrant.
  1, 5, and 2 cases are out of bounds (>\,40\%) in German, Dutch, and Italian, respectively.}
  \label{figure:indiv-analysis}
 \end{figure*}

We look at the performance changes in individual cases.
Figure \ref{figure:indiv-analysis} shows scatter diagrams of the four languages.
In these diagrams, we plot the averaged F1-score changes of 15 runs by adding (i.e., from \textsc{no char} to \textsc{unigrams}) and extending (i.e., from \textsc{unigrams} to \textsc{bigrams}) character-level information.
We observe many cases whose averaged F1-score increases with the addition and extension of character-level information (plotted in the first quadrant).
However, these numbers are lower than those in the second and fourth quadrants, indicating that the improvement works only by either adding or extending the information.
Moreover, we observed cases whose scores decrease in both aspects, plotted in the third quadrant.
These trends are observed for all languages, even though the overall scores improved for all languages except English.

\subsection{Why do our values deviate from \vannoord{}?}
\label{section:vannoord-german}
The values reported in this study are lower than those from the previous study \vannoord{}, especially in German.
We follow nearly all the setups reported in \vannoord{}, but the values are still low.

\Vannoord{} reports that they only used the gold and silver data if gold (train) data is available in a certain language.
The German data in PMB release 3.0.0 has the gold train data comprising 1,159 documents.
Therefore, we experiment with the model pre-trained on the merged set of the gold and silver data and fine-tuned on the gold data only.
We reported an averaged value of five runs in Table \ref{table:de} with one from \vannoord{}.
A large deviation between the two F1-scores can be observed.

\begin{table*}[t]
  \centering\small
  \begin{tabular}{lcc} \toprule
    & Average          & All values \\ \midrule
\Vannoord{} & 82.0~\,\, & N/A \\
Our replication & 68.52 & 68.54,\, 67.95,\, 69.38,\, 68.61,\, 68.10 \\ \bottomrule
\end{tabular}
  \caption{F1-scores (\%) from \vannoord{} and our replication experiment in German.
  The models is pre-trained on the unified set of the gold and silver train data and fine-tuned on the gold train data.}
  \label{table:de}
\end{table*}

\end{document}